\documentclass{esannV2}
\usepackage[dvips]{graphicx}
\usepackage[latin1]{inputenc}
\usepackage{amssymb,amsmath,array}
\usepackage{adjustbox}

\begin{document}
\title{A simple technique for improving multi-class classification with neural networks}

\author{Thomas Kopinski$^1$, Alexander Gepperth$^2$ and Uwe Handmann$^1$
%
%
\vspace{.3cm}\\
%
1- University Ruhr West - Computer Science Institute \\
L\"utzowstrasse 5, 46236 Bottrop - Germany
%
\vspace{.1cm}\\
2- ENSTA ParisTech\\
828 Blvd des Mar\'echaux, 91762 Palaiseau, France
}

\maketitle

\begin{abstract}
We present a novel method to perform multi-class pattern classification with neural networks and test it on a challenging 3D hand gesture recognition problem. Our method consists of a standard one-against-all (OAA) classification, followed by another network layer classifying the resulting class scores, possibly augmented by the original raw input vector. This allows the network to disambiguate hard-to-separate classes as the distribution of class scores carries considerable information as well, and is in fact often used for assessing the confidence of a decision. We show that by this approach we are able to significantly boost our results, overall as well as for particular difficult cases, on the hard 10-class gesture classification task. 
\end{abstract}
\section{Introduction}
The problem of multi-class pattern classification can be defined as follows: For a sample $X_i$ taken from a set of n samples $X_1$,...,$X_n$ where each sample is characterized by m features $a_1$,...,$a_m$ $\in$ A find a function $f:X\rightarrow L$ which maps the sample to the correct corresponding label $l_i$ $\in$ $L$ : $\{$ $l_1$,...,$l_n$ $\}$. This is contrasted to the binary classification problem where $l$ belongs to either class 0 or 1 (or -1,1) respectively. The generalization performance of such a function measures how well it is able to predict the correct label for a previously unseen sample.\\
Various approaches exist for this problem and one generally distinguishes one-against-one (OAO), one-against-all(OAA) and P-against-Q (PAQ) methods (see \cite{ou2007multi}). Most of these methods are rather empirical in nature, and a precise mathematical theory to guide our choice of algorithms, as it exists for binary classification, seems unavailable. We present a novel approach which is somewhere in the middle between OAA and OAO, applied to the problem of hand pose recognition with a single ToF\cite{kopinskineural}. The basic idea is to build a two-stage architecture that first performs a normal OAA classification, and subsequently fuses the obtained set of class scores with the original network input for a second layer of classification. To verify our approach, we create a very large a database of training and testing samples which is divided accordingly to the needs of the problem, and are able to show that the proposed approach can improve the overall generalization performance, as well as significantly boosting performance on specific difficult classes.\\
%
\subsection{Related work}
The problem of multi-class classification has been thoroughly researched in \cite{ou2007multi} with special emphasis of modelling pattern classes with one-against-all, one-against-one and P-against-Q.  This study compares the advantages and disadvantages of each of the approaches by employing either a single or a series of neural networks supported by various decision functions. The results of this analysis show the benefit and drawbacks of splitting the task into multiple independently trained neural networks of which some are the uncovered feature space regions and ambiguity issues. The authors of \cite{murphey2007oaho} show the increased difficulty in the classification problem with an imbalanced class distribution. In order to solve the multi-class classification problem with support vector machines without creating independent classifiers a single optimisation routine has been proposed in \cite{weston1999support}. Whether to rely on an OAO, OAA or an PAQ approach is a non-trivial question and a more in-depth analysis in favor of the OAA approach can be found in \cite{rifkin2004defense}. The most important statement made is that once the best binary classifier has been chosen the choice of the multiclass scheme becomes secondary and the simpler scheme should be preferrable. In \cite{ding2001multi} the authors successfully show how a multi-class classifier can be efficiently applied to the problem of protein folding with a special focus on the comparison of various discriminative methods. They achieve impressive results by employing support vector machines as classifiers. In \cite{stallkamp2011german} the performance of different neural network topologies is tested on the multi-class problem of traffic sign recognition.  
\section{Database and Descriptors}
\begin{figure}
\includegraphics[width=1.0\textwidth]{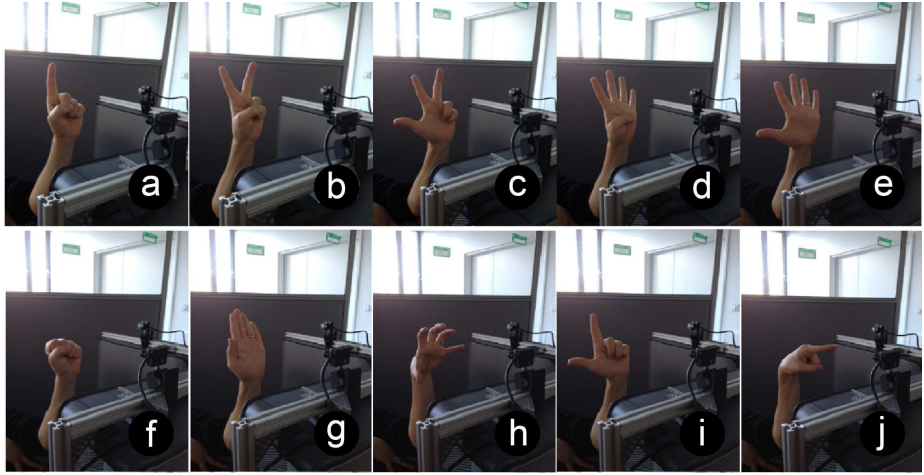}
\caption{The 10 different gestures recorded from 15 different persons}
\label{fig:gestures}
\end{figure}

We recorded a database containing 3000 samples for each of 10 gestures collected from 15 persons, see Fig.\ref{fig:gestures}. To increase variance in the data the snapshots were taken in three regions in a range from 20cm-50cm and the participants were asked to translate and rotate their hands during the recording. Overall, the total sample count is 450000 samples.
Each sample originates from raw point cloud data which is transformed via a global Point Cloud Descriptor into a histogram of fixed size. The descriptor depicts the global shape of the cloud via the relationship of the angles calculated from a sample subset of point pairs. For a more in-depth specification of this feature transformation please refer to \cite{kopinskineural}. Since we are interested in the multi-class classification task of recognizing shapes, we automatically remove the background from the recording by simple distance thresholding. The gestures chosen for this classification task are relevant to HMI applications and contain a significant amount of ambiguity, thereby creating a hard problem for a meaningful benchmark.
\section{Multi-class classification and fusion technique}
\subsection{Neural network topology}
We aim for a OAA classification approach and thus design our neural networks accordingly. 
We implement a 2-step approach involving two distinct multilayer perceptron (MLP) networks.
Each neural network has three layers - input, hidden and output layer with the input and hidden layers fully connected to their subsequent layers. The input for the first net is formed by the point cloud descriptor of the current sample, the output layer consists of 10 output neurons, one for each class. The input for the second net differs as it not only takes into account the descriptor of the current sample but also the output values of the first net which classifies the same sample. The determination of the number of hidden neurons for each net is described in the following chapter. The output layer of the second network also consists of 10 neurons.

\subsection{Early fusion technique}

In order to train the two-stage architecture described in the previous chapter, 
the available data samples need to be divided into three disjunct sets. 
We distinguish between the training (D1 and D2) and generalization (D3) sets, the latter consisting of all the samples (about 30000) coming from person $1\leq i \leq 15$. The remaining samples are divided into the two equally sized sets D1 and D2. In order to monitor training performance we furthermore randomly split each D1 and D2 in a ratio 9:1 into the sets D1.train, D1.test, D2.train and D2.test. These sets are used during the training of each network to achieve a more independent measurement of the current generalization performance. For the first network, D1.train is used for training the network while the mean squared error measured on the D1.test set is used for selecting the best-performing net among all conducted iterations. As soon as the first net is trained, the second net is trained on D2 in an analogous manner. The main difference is that the input for the second net is formed as follows: For a sample taken from D2.train, feed this sample into the first net and memorize the output of all 10 output neurons. These values are concatenated with the descriptor (size $n$) of the sample, thus having length $n$ + 10, and forms the new training sample fed into net 2, with supervision coming again from the known class of the sample. The rest of the procedure remains the same, D2.test is used as the performance measure for the decision process of the second phase of training.
\section{Experiments and Results}
All experiments were conducted with the FANN library and implemented under Ubuntu in C++. We have tested several configurations in order to determine the hidden layer size and optimal training parameters. Initial test runs have shown that the best generalization results can be achieved when setting the size of the hidden layer between 25-40 neurons for both networks. The remaining parameters were chosen as standard values and testing various configurations provided no meaningful insight regarding our problem. The training algorithm is RPROP, the activation function is the symmetric sigmoid function for both hidden and output layers.\\
On average we were able to improve the generalization results by between 2-4\% considering all classes, as Table \ref{tab:res1} shows, when employing two nets each having 40 neurons in the hidden layer. The upper row shows the results for net 1 alone, being fed the feature vector only while the lower row displays the fused feature and neuron values. In 3 cases we see a slight drop in generalization performance, in one single case no change is notable while all the remaining cases show partly significant improvements when fusing the data with output activities. In some selected cases we have improvement rates of 5-9\% averaged over all the gestures of one person.

\begin{table}

\resizebox{\textwidth}{!}{\begin{tabular}{|l|l|l|l|l|l|l|l|l|l|l|l|l|l|l|}

\hline
1 & 2 & 3 & 4 & 5 & 6 & 7 & 8 & 9 & 10 & 11 & 12 & 13 & 14 & 15 \\ \hline
.81 & .43 & .68 & .43 & .83 & .49 & .53 & .68 & .81 & .70 & .88 & .73 & .95 & .66 & .76\\ \hline
.83 & .44 & .67 & .54 & .84 & .53 & .55 & .67 & .78 & .70 & .91 & .74 & .97 & .73 & .90\\ \hline

\end{tabular}}
\caption{Comparison of the generalization performance of net 1 (upper row, 40 hidden neurons) vs. net 2 (lower row, 40 hidden neurons). The indexed columns indicate the generalization data coming from person i.}
\label{tab:res1}	

\end{table}

As a comparison Table \ref{tab:res2} displays the performance of both nets each with 25 and 20 hidden neurons respectively. The main difference in this case is the fact that the nets with smaller hidden layer seem to perform slightly more poorly while we are still able to show an improvement with our fusion technique when individually comparing each generalization case. Moreover the most significant improvements can be found in both experiments.

\begin{table}
\resizebox{\textwidth}{!}{\begin{tabular}{|l|l|l|l|l|l|l|l|l|l|l|l|l|l|l|}
\hline
1 & 2 & 3 & 4 & 5 & 6 & 7 & 8 & 9 & 10 & 11 & 12 & 13 & 14 & 15 \\ \hline
.81 & .40 & .68 & .39 & .81 & .48 & .52 & .68 & .79 & .68 & .88 & .71 & .94 & .64 & .74 \\ \hline
.81 & .45 & .67 & .50 & .83 & .53 & .52 & .67 & .77 & .67 & .90 & .71 & .96 & .68 & .89 \\ \hline
\end{tabular}}

\caption{Comparison of the generalization performance of net 1 (upper row, 25 hidden neurons) vs. net 2 (lower row, 20 hidden neurons)}
\label{tab:res2}
\end{table}

Table \ref{tab:res3} shows the change in recognition rate in reference to the individual hand gesture classes averaged over all persons. In 8 of 10 cases we achieve a boosting of the recognition rate except for gesture classes c and d (-2.1\% and -2.8\%). The best improvement amounts to 4.7\% and 4.8\% which is, for this problem, of special interest since these are two of the more difficult cases to disambiguate (cf. Fig. \ref{fig:gestures}).

\begin{table}[h]
\begin{tabular}{|l|l|l|l|l|l|l|l|l|l|}
\hline
a & b & c & d & e & f & g & h & i & j \\ \hline
+1.3 & +4.8 & -2.1 & -2.8 & +3.8 & +2.0 & +3.1 & +3.2 & +4.7 & +2.5 \\ \hline
\end{tabular}
\caption{The average increase/decrease in recognition rate for all 10 gestures averaged over all the persons for net 1 compared to net 2 (both 40 hidden neurons).}
\label{tab:res3}
\end{table}
Table \ref{tab:res4} shows the corresponding improvements for the recognition rate averaged over all persons for all individual gesture classes. 
\begin{table}[h]
\begin{tabular}{|l|l|l|l|l|l|l|l|l|l|}
\hline
a & b & c & d & e & f & g & h & i & j \\ \hline
+.01 & +.02 & +.01 & -.03 & +.07 & +.01 & +.05 & +.04 & +.01 & +.04 \\ \hline
\end{tabular}
\caption{The average increase/decrease in recognition rate for all 10 gestures averaged over all the persons for net 1 compared to net 2 (25 and 20 hidden neurons respectively).}
\label{tab:res4}
\end{table}
We equally test our approach by randomly splitting the whole dataset (i.e., not taking into account individual persons) into two equally sized datasets, one for training and one for evaluation. Since the sets were not completely independent, the overall classification rate was 86\% for net 1 and 87\% for net 2 which is not surprising but still shows we are able to improve the results with our fusion approach.  
A comparison of our approach with the MNIST dataset yielded, for 15 different test runs conducted in the same manner (as there is no person information available), neither a significant drop nor an improvement since the classification error remained around 5.5\% for net 1 and net 2 with a slight variance. This variance can be ascribed to the random initialization of the weights. It should be noted that we split our dataset into three parts namely 40\% for training each net and 20\% for testing, which is a significantly reduced training data set.   
\section{Discussion and Outlook}
This contribution presents a novel fusion technique for the multi-class pattern classification problem of recognizing hand poses with neural networks. The described approach is a 2-stage deep neural network with a total of 5 layers. Our fusion approach exceeds the performance of a regular neural network by about 2-3\% measured as the classification errors averaged over 15 persons, each of which was left out from the training set, amounting to a sort of cross-validation procedure. More specifically we are able to significantly boost the results in a number of cases, especially for poses that are more difficult to disambiguate. Nevertheless in two cases we see a decrease in classification performance which has also to be taken into consideration when designing a real-time applicable system. In these cases however one can simply rely on the first net when designing a decision function and use the second net for the other instances.\\
In terms of applicability we have implemented this fusion technique into our real-time hand gesture recognition system and it has shown to stabilize the results when designing the decision process as described above \cite{kopinski2014real}. We are confident to be able to improve our performance even further by employing confidence measures for rejecting false positives and thus stabilizing the recognition performance. 
%

\begin{footnotesize}


\bibliographystyle{unsrt}
\bibliography{bibfile}

\end{footnotesize}


\end{document}